\documentclass[letterpaper, 10 pt, conference]{ieeeconf}  % Comment this line out if you need a4paper

\IEEEoverridecommandlockouts                              % This command is only needed if 
\usepackage{times}
\usepackage{multicol}
\usepackage[bookmarks=true]{hyperref}
\usepackage{amsfonts}
\usepackage{amsmath,bm}
\usepackage{amssymb}
\usepackage{CJK}
\usepackage{indentfirst}
\usepackage{cases}
\usepackage{algorithm}
\usepackage[algo2e]{algorithm2e}
\usepackage{graphicx}
\usepackage{stfloats}
\usepackage{subfigure}
\usepackage{booktabs}
\usepackage{color,soul} %\st
\definecolor{red}{rgb}{1.00,0.00,0.00}
\definecolor{blue}{rgb}{0.00,0.00,1.00}
\definecolor{green}{rgb}{0.30, 0.50,0.00}

\newcommand{\cblue}[1] {\textcolor{blue}{#1}}

\usepackage{flushend} % aligned references columns
\usepackage{arydshln}
\usepackage{color}
\usepackage{tikz}
\usepackage{cite}

\usepackage{pifont}
\usepackage{multirow}

\title{\LARGE \bf Simulation-Driven Evolutionary Motion Parameterization for Contact-Rich Granular Scooping with a Soft Conical Robotic Hand}

 \author{Yongliang Wang$^{1, 2}$,  Cristian C. Beltran-Hernandez$^{*,1}$, Tomoya Takahashi$^{1}$,  and Masashi Hamaya$^{1}$% <-this % stops a space
 \thanks{This work was supported by the JST-Mirai Program Grant Number JPMJMI21G2, Japan.}% <-this % stops a space
 \thanks{$^{1}$ OMRON SINIC X Corporation, Tokyo, Japan.}
 \thanks{$^{2}$ Department of Artificial Intelligence, Bernoulli Institute, Faculty of Science and Engineering, University of Groningen, The Netherlands}
 \thanks{* Corresponding author:}
 \thanks{\tt\small   ~~~cristian.beltran [at] sinicx.com}%
 }

\begin{document}

\maketitle
\thispagestyle{empty}
\pagestyle{empty}

%%%%%%%%%%%%%%%%%%%%%%%%%%%%%%%%%%%%%%%%%%%%%%%%%%%%%%%%%%%%%%%%%%%%%%%%%%%%%%%%
\begin{abstract}

Tool-based scooping is vital in robot-assisted tasks, enabling interaction with objects of varying sizes, shapes, and material states. 
Recent studies have shown that flexible, reconfigurable soft robotic end-effectors can adapt their shape to maintain consistent contact with container surfaces during scooping, improving efficiency compared to rigid tools. These soft tools can adjust to varying container sizes and materials without requiring complex sensing or control. However, the inherent compliance and complex deformation behavior of soft robotics introduce significant control complexity that limits practical applications. To address this challenge, this paper presents the development of a physics-based simulation model of a deformable soft conical robotic hand that captures its passive reconfiguration dynamics and enables systematic trajectory optimization for scooping tasks. We propose a novel physics-based simulation approach that accurately models the soft tool's morphing behavior from flat sheets to adaptive conical structures, combined with an evolutionary strategy framework that automatically optimizes scooping trajectories without manual parameter tuning.
We validate the optimized trajectories through both simulation and real-robot experiments. The results demonstrate strong generalization and successfully address a range of challenging tasks previously beyond the reach of existing approaches. Videos of our experiments are available online: \href{https://sites.google.com/view/scoopsh}{\cblue{https://sites.google.com/view/scoopsh}}
% TODO: add project page and code 

\end{abstract}

%%%%%%%%%%%%%%%%%%%%%%%%%%%%%%%%%%%%%%%%%%%%%%%%%%%%%%%%%%%%%%%%%%%%%%%%%%%%%%%%

\section{Introduction}
\label{sec:introduction}

Scooping is an instinctive and essential human skill, allowing us to efficiently handle materials such as fluids and granules in tasks ranging from ladling soup and collecting peas at the dining table to excavating soil at construction sites \cite{tai2023scone, mehta2025l2d2, wang2021learning, schenck2017learning, ruan2024primp, song2025soda}. In robotics, scooping has the potential to greatly benefit daily life, particularly in food preparation and assistive feeding across homes, hospitals, and restaurants \cite{keely2025kiri}. However, despite advances in robotics, stable and adaptive autonomous scooping has received limited attention. The task remains challenging due to the complex interactions between the end-effector and dynamic materials. Most robotic scooping tools are rigid and fixed in shape, lacking the passive compliance needed for smooth, precise motion across diverse containers \cite{zhao2025learning}. By contrast, humans naturally adapt their choice of tools to container geometry and fill level, an adaptability that robotic systems still struggle to achieve.

\begin{figure}[!t]
      \centering
      \includegraphics[width=1.0\linewidth]{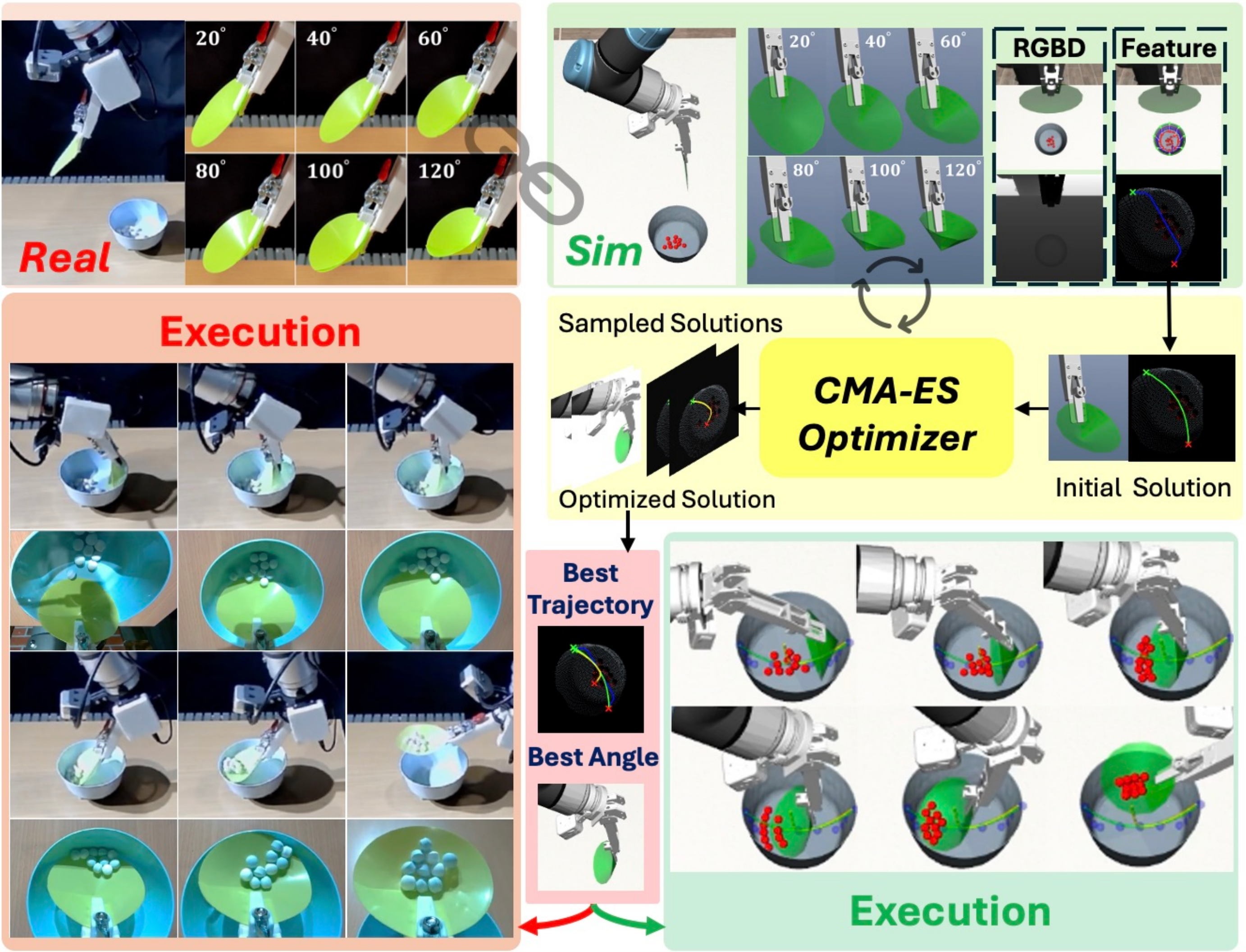}
      \vspace{-4mm}
      \caption{\textbf{Contact-rich granular scooping system:} The framework couples real-world execution (left) with its digital twin in simulation (right). From RGB-D perception and feature extraction, container geometry is abstracted into seed trajectories. In simulation, the covariance matrix adaptation evolution strategy (CMA-ES) optimizes both trajectory and hand roll angle, producing the best solution. This optimized strategy is then directly transferred to the real soft hand without additional tuning, as demonstrated in the 10-ball scooping task.}
      \label{fig: overview}
\end{figure}

Although scooping trajectories may appear similar at first glance, achieving successful execution is challenging under different environmental settings. Current research on robotic scooping mainly addresses three aspects of environment design: tools, objects, and containers \cite{franco2024double, liu2024diff}. 
% Reviewer 1 comment regarding redunancy
% CB: removed most references as they will be introduced later.
\textbf{Tools:} Most studies employ rigid spoons or other fixed-shape utensils designed primarily for food handling, with only a few works exploring deformable tools tailored to specific assistive-feeding scenarios. \textbf{Objects:} Scooped materials range from solid particles to granular media and fluids, but in many trajectory-generation approaches material properties play a secondary role compared to geometry and contact. \textbf{Containers:} Experimental setups are often simplified to open boxes or regular bowls, whereas narrow, deep, or irregular containers pose substantially greater challenges and remain underexplored.
In summary, completing a scooping task requires answering three questions: \textit{What tool is used? What material is being scooped? Where is the operation performed?} In this paper, we primarily focus on the first and third aspects.

For generating scooping trajectories, many scooping methods rely on geometric representations. Some assume a standard container with predefined trajectories, later generalized \cite{das2024screw}, while others use vision to extract features and map them to handcrafted motion primitives. Though robust, these approaches are hand-engineered and inflexible. Learning-based methods \cite{yang2025s} offer adaptability but demand large amounts of real or human-collected data. Limited work addresses simulation, as modeling physics, tools, and material interactions is difficult. Thus, building an end-to-end framework from simulation to trajectory generation remains an open challenge. From prior work, three key limitations can be identified: (1) Most approaches assume fixed environment settings, for example, a spoon of fixed size, a container of fixed geometry, and a fixed quantity of material, making generalization to varied environments difficult; (2) Existing tools are not designed for adaptability, even though real-world scooping tasks require handling diverse materials and containers; and (3) The lack of accurate simulation models for both tools and materials hinders efficient data collection for optimization and learning. A recent study \cite{takahashi2025scu} has proposed a tool with the potential to address the first two limitations, but the challenge of realistic simulation remains. Our work primarily focuses on tackling this remaining gap.

To address these limitations, particularly the challenge of building accurate models in simulation, we propose a complete framework that integrates the design of a soft deformable gripper with its simulation model in MuJoCo (see Fig.~\ref{fig: overview}). This modeling approach leverages the unique properties of MuJoCo and can serve as inspiration for simulating other soft robotic tools. Furthermore, we map scooping actions from visual information and employ an evolution-based optimization system to efficiently explore the trajectory parameter space, converging to locally optimal parameters that maximize scooping success. In our experiments, we first optimize scooping parameters in simulation and then validate the resulting trajectories on a real robot, demonstrating minimal sim-to-real gap. The primary contributions of this paper are summarized as follows:

\begin{itemize}
    \item We develop a physics-based MuJoCo simulation model of a soft deformable hand for scooping and validate it by optimizing motion parameters in simulation and transferring directly to a real robot, achieving reliable scooping with minimal sim-to-real gap.
    \item We propose a framework that maps visual information to primitive scooping motions. Raw trajectories are generated from features of the environment, such as container shape and size, and then optimized using an evolutionary strategy to improve scooping success.
\end{itemize}

\section{Related work}
\label{sec:related_work}

\subsection{Robotic Scooping Strategies}
\label{subsec:robotic_scooping_strategies}
Recent research has explored a wide range of robotic scooping methods for handling diverse materials. Successful scooping depends on several key factors, the scooping task can be broadly divided into three aspects: \textbf{tool design and utilization}, which examines different scooping implements and their suitability for specific applications; \textbf{object properties}, which consider how material type, shape, and physical state affect scooping performance; and \textbf{motion generation strategies}, which focus on how robots plan and execute effective scooping trajectories.

\subsubsection{Robotic Scooping Tools}
A spoon is the most common tool for scooping. In robotics research, small rigid spoons are typically employed, especially in feeding-related studies. For example, \cite{wang2021learning} demonstrated a real-world robot executing trained scooping primitives in KitchenPR2 across varying bowls, spoons, and spoon poses, while \cite{mehta2025l2d2} introduced L2D2, where humans taught robots to scoop by sketching trajectories on workspace images. A broad range of works \cite{kageyama2024learning, bhaskar2024lava, ruan2024primp, yang2025s} further explored spoon-based scooping in diverse contexts, including contact-rich tasks, granular media, and deformable substances, highlighting the central role of spoons in robotic scooping research. In addition to small spoons, larger soup spoons have been investigated. \cite{tai2023scone} used a stainless steel soup spoon in food scooping tasks, while \cite{lin2025robotsmith} showcased spoon-based scooping within a tool-design framework. Other studies extended spoon usage to granular and liquid media: \cite{schenck2017learning} attached a rigid spoon to manipulate granular substances, \cite{liu2024diff} adopted a large soup spoon for water-based scooping of floating objects, and \cite{moorman2024investigating} equipped a robot with a soup spoon to mix soil components. Beyond spoons, other rigid tools were also explored. \cite{niu2023goats} studied water scooping with bowls and buckets, \cite{franco2024double} introduced a tendon-driven Gripper with scoop-shaped fingertips. Soft tools for scooping also attracted growing interest, though such designs remained rare due to their mechanical complexity. \cite{song2025soda} presented a 3D-printed utensil with origami-inspired artificial muscles that flexibly switched between gripping and scooping for different food textures. Similarly, \cite{keely2025kiri} introduced a kirigami-based soft spoon capable of deforming into a bowl to wrap, contain, and release food during robot-assisted feeding. In summary, while robotic scooping has been studied with a variety of tools, most works focus on rigid spoons and devote little attention to tool design itself. Few approaches attempt to combine rigidity, softness, and deformability to achieve versatile and practical scooping across diverse tasks. Our work seeks to address these limitations.

\subsubsection{Scooped Objects}
Research on robotic scooping can also be categorized by the properties of the target objects. Many studies focus on small solid particles. For example, \cite{tai2023scone} trained on food items such as rice, beans, and chocolate balls of varying sizes, shapes, and weights, demonstrating strong generalization to unseen items. Similarly, rice was used in \cite{das2024screw}, while cereals were considered in \cite{mehta2025l2d2, wang2021learning, yang2025s}. Other works extended to larger discrete items like beans \cite{yang2025s, wang2021learning}, granular media such as sand \cite{schenck2017learning}, and even fine powders \cite{ruan2024primp}. Beyond rigid particles, some research addressed complex food materials with sticky or hard-to-cut properties \cite{song2025soda, franco2024double}. Fluid scooping had also been investigated, including water \cite{niu2023goats} and sauces \cite{lin2025robotsmith}. Applications extended further to soil, stones, and mining, as in autonomous excavation \cite{franceschini2024autonomous}, as well as deformable plastic-like substances \cite{kageyama2024learning}. 

\subsubsection{Scooping Motion Generation}
Research on robotic scooping spans geometric, learning-based, and interactive approaches. Geometric methods construct trajectories from shape-based representations \cite{das2024screw}, while general motion generation targets robustness across varied terrains \cite{franceschini2024autonomous}. Data-driven methods dominate: imitation learning from demonstrations enables category-level generalization \cite{yang2025s}, long-horizon food acquisition \cite{bhaskar2024lava}, motion primitives \cite{ruan2024primp}, and sketch-based inputs \cite{mehta2025l2d2}. Reinforcement learning refines strategies via trial and error \cite{niu2023goats}, and behavior cloning has been applied to food scooping \cite{tai2023scone}. Advanced architectures extend these directions, including VLMs for tool design \cite{lin2025robotsmith}, diffusion policies for scooping \cite{yang2025s}, and transformers for tactile perception and granular media modeling \cite{kageyama2024learning,schenck2017learning,wang2021learning}.
Our proposed physics-based simulation environment can greatly benefit existing approaches, particularly those that are data-intensive and require substantial demonstration data, by providing a realistic and cost-effective platform for trajectory generation and optimization.

\subsection{Soft and Deformable Tools in Robotic Manipulation}
\label{subsec:soft_tools}
Rigid tools have clear limitations when handling diverse or delicate scooping tasks, motivating researchers to investigate soft and hybrid alternatives. For example, \cite{zhu2022soft} proposed a soft–rigid hybrid gripper actuated by pneumatic muscles, enabling dexterous multi-DOF grasping. Similarly, \cite{mehta2023riso} introduced grippers that combined rigid fingers with switchable soft adhesives. Extending this multi-modal functionality, \cite{wang2025dexgrip} presented a soft robotic gripper with an active suction palm and rotating grasping surfaces. Bio-inspired approaches further broadened the design space. \cite{guo2025fish} introduced a fish-mouth-inspired origami gripper for underwater grasping and scooping, demonstrating robustness in handling marine creatures. However, when faced with containers of varying shapes and sizes, even soft or hybrid tools often failed to complete scooping tasks. To address these challenges, researchers began exploring deformable tools. \cite{huang2025dih} presented a teleoperation framework combining a tactile dexterous hand with multimodal perception to achieve complex in-hand manipulations. In parallel, \cite{zhao2025learning} proposed a soft underactuated hand with passive compliance and integrated sensing, and successfully demonstrated tasks such as fabric display and page turning. \cite{takahashi2025scu} introduced a flexible reconfigurable end-effector for powder scooping, which we adopt as the tool in this paper. These works highlight the potential of deformable tools to extend robotic manipulation capabilities beyond the limits of rigid and soft designs. 

\subsection{Simulation for Soft Robotic Systems}
Simulation is fundamental for advancing soft robotics, enabling controlled experimentation, design exploration, and policy learning. Differentiable simulation has been proposed as a means to jointly optimize robot design and control through gradient-based methods \cite{bacher2021design}. To support evaluation, benchmarks such as SoftGym \cite{lin2021softgym} provide standardized tasks for deformable object manipulation. Platforms like Elastica \cite{naughton2021elastica} extend simulation to continuum arms. Sim-to-real approaches have also been explored in the context of soft robotic wrists for insertion tasks \cite{fuchioka2024robotic}. In tactile sensing, finite element simulation has been applied to soft-bubble sensors for contact and force estimation \cite{peng20243d}, while differentiable methods have been developed for efficient gel-based surface tactile simulation \cite{xu2023efficient}. Origami simulators based on rigid-panel models \cite{zhu2022review, tachi2009simulation} capture folding well and inspire origami/kirigami robotic tools, but mainly serve for visualization. Our work instead integrates deformable tool simulation into a contact-rich scooping framework, enabling optimization and sim-to-real transfer. Together, these works establish a foundation for developing simulators tailored to soft robotic systems. However, very few studies have developed soft, deformable tools specifically for scooping. Such tools are also uncommon in real-world applications, and accurately modeling them in simulation remains particularly challenging. The absence of robust simulation models limits the use of advanced learning methods, constraining efforts to explore, validate, and scale deformable-tool-based scooping.

\begin{figure}[!t]
      \centering
      \includegraphics[width=1.0\linewidth]{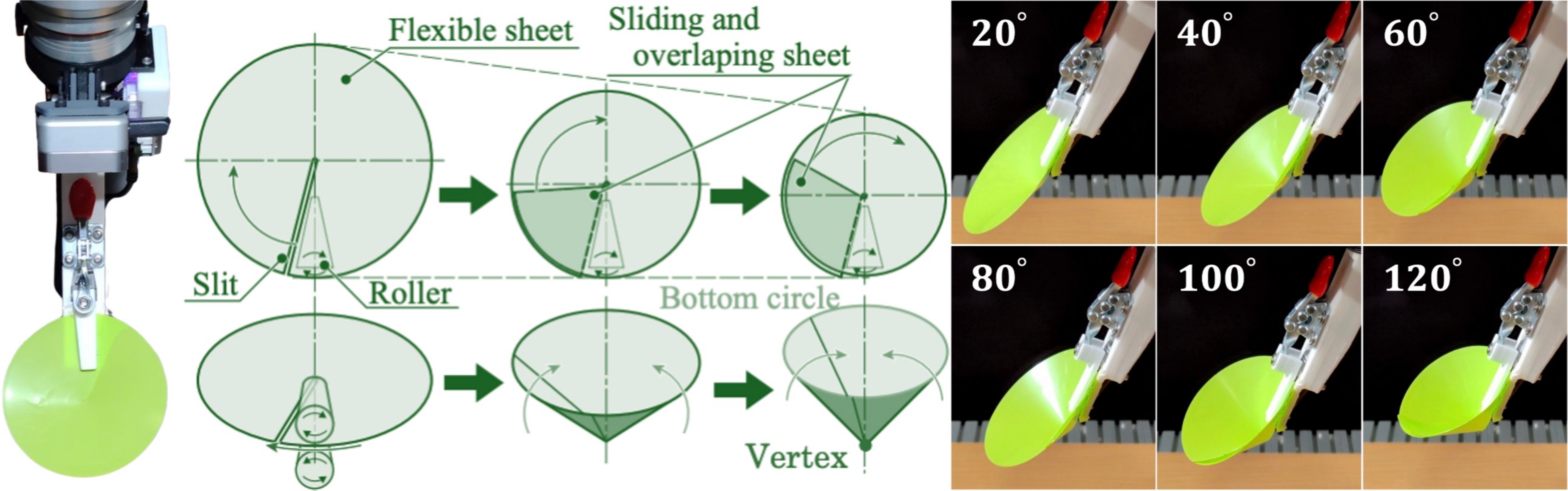}

      \caption{The real soft-hand \cite{takahashi2025scu} rolling from $20^\circ$ to $120^\circ$ in $20^\circ$ increments.}
      \label{fig: real_scu}
\end{figure}

\section{Method}
\label{sec:method}

\subsection{System Overview}
\label{subsec:sys}
While novel soft and deformable tools have been developed, building simulation models for manipulation tasks remains challenging, alongside the difficulty of collecting large-scale real-world data. 
This study employs the soft-hand (SH) mechanism originally proposed in \cite{takahashi2025scu}, shown in Fig.~\ref{fig: real_scu}. 
The mechanism is modeled in MuJoCo, where an RGB-D camera captures the environment and FastSAM segments the target container \cite{zhao2023fast}. Segmentation results form the basis for a raw scooping trajectory, which is subsequently optimized in simulation via CMA-ES, iteratively refining motion parameters according to interaction dynamics. The trajectories optimized in simulation are directly transferred to the physical robot for validation of real-world effectiveness.

\subsection{Soft Hand: Mechanical Design and Simulation}
\label{subsec:soft_hand_design_and_simulation}

\subsubsection{Mechanical Design of the Soft Hand}
The SH incorporates a flexible, conical structure that conforms to various container geometries through passive deformation, ensuring consistent contact without the need for complex force sensing or machine learning–based control strategies. Its reconfigurable mechanism further allows size adjustment, enabling efficient scooping across a wide range of container types. The SH is constructed from a circular, flexible sheet, a supporting frame, rollers, a drive shaft, and a motor. The prototype is mounted at a $45^\circ$ angle relative to the manipulator to facilitate insertion into deep containers. The end-effector is offset $150 mm$ from the motor to extend its reach. A $0.2 mm$ thick polypropylene sheet (initial diameter $100 mm$) is used as the flexible tool. The roller assembly consists of a driven nitrile rubber roller and a free polyacetal roller, which together pinch the sheet to generate motion (Fig.~\ref{fig: real_scu}).

\subsubsection{MuJoCo Simulation Model}

To replicate the behavior of the SH hand in simulation, we employ the MuJoCo physics engine, which provides physics-based contact dynamics, continuous-time physics, and support for deformable structures \cite{todorov2012mujoco, zakka2025mujoco}. A central challenge in this work lies in modeling a soft, funnel-like sheet that can deform realistically under roller actuation. While MuJoCo provides examples of deformable objects such as jelly, cloth, or fluid, these demonstrations are typically passive and lack mechanisms for precise, task-oriented control. In contrast to rigid-body approximations commonly employed in dexterous hand simulation, our approach combines a custom-generated geometric mesh with a staged, constraint-based control strategy. This integration enables both realistic passive compliance and controllable deformation. To the best of our knowledge, this is the first method to achieve such a controllable deformable-hand model within the MuJoCo simulator.

\begin{figure}[!t]
      \centering
      \includegraphics[width=1.0\linewidth]{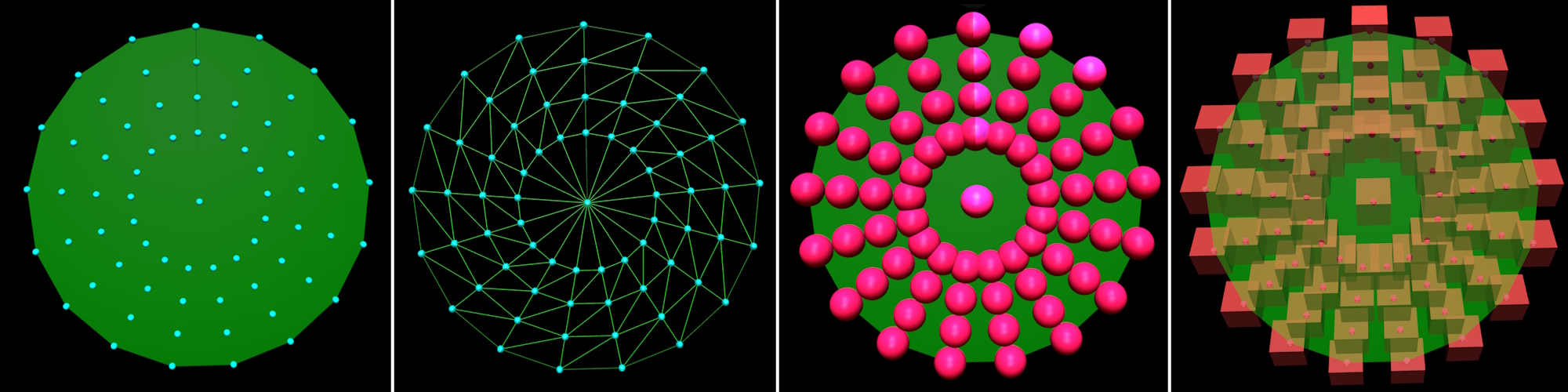}

      \includegraphics[width=1.0\linewidth]{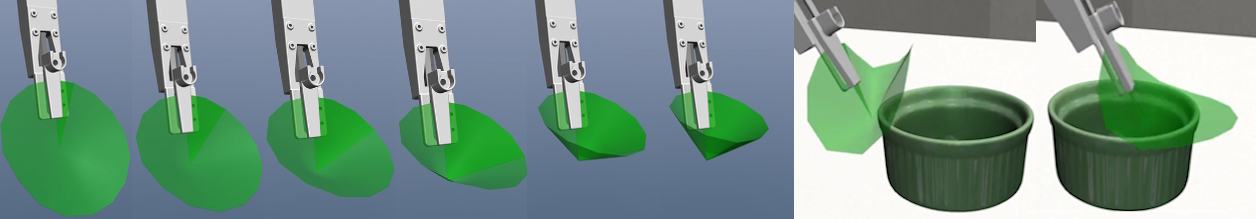}

      \caption{The top row shows, from left to right, the Model View, Wireframe View, Mesh View, and Inertia View. The bottom row illustrates the hand rolling from $20^\circ$ to $120^\circ$ in $20^\circ$ increments, highlighting its soft and deformable properties.}
      \label{fig: sim_scu}
\end{figure}

\paragraph{Geometry Construction}  
The top row of Fig.~\ref{fig: sim_scu} shows how the flexible sheet was represented in simulation. The sheet is modeled as a network of connected triangles arranged in rings, using the \texttt{trimesh} tool. A section of the sheet is cut away to match the real scooping tool, which needs an opening to create a funnel shape when scooping. By adjusting the number of rings, the spacing between points, and the size of the cutout, the simulation strikes a balance between precision and stability. Seventy-two points on the sheet’s face are matched with small spheres in the simulation, and these spheres are locked to their corresponding sheet points with constraints (equality \emph{weld}) that prevent movement between them. This approach ensures that simulated contact, like the tool touching container surfaces, happens exactly at the edge of the sheet, while the sheet itself can bend and flex realistically.

\paragraph{Dynamics and Parameterization}  
The SH sheet is modeled as a mass–spring–damping system, where each mesh node obeys
\begin{equation}
    m_i \ddot{\mathbf{x}}_i = \mathbf{f}_i^{\mathrm{int}} + \mathbf{f}_i^{\mathrm{ext}} - c \dot{\mathbf{x}}_i
\end{equation}
where $m_i$ is the mass associated with node $i$, $\mathbf{x}_i$ is the position vector of node $i$, 
$\ddot{\mathbf{x}}_i$ and $\dot{\mathbf{x}}_i$ are its acceleration and velocity, respectively, 
$\mathbf{f}_i^{\mathrm{int}}$ denotes the internal elastic force on node $i$ arising from stretching and bending, 
$\mathbf{f}_i^{\mathrm{ext}}$ is the external force applied to node $i$ (e.g., gravity or contact with the container), 
and $c$ is the Rayleigh damping coefficient. The elastic energy of the sheet is given by
\begin{equation}
    E = \tfrac{1}{2} \sum_{(i,j)} k_s \big(\lVert \mathbf{x}_i - \mathbf{x}_j \rVert - l_{ij}^0 \big)^2 
  + \tfrac{1}{2} \sum_{(p,q)} k_b \, (\theta_{pq} - \theta_{pq}^0)^2
\end{equation}
where $\mathbf{x}_i$ and $\mathbf{x}_j$ are the positions of nodes $i$ and $j$,  
$k_s$ and $k_b$ denote the stretching and bending stiffness coefficients,  
$l_{ij}^0$ is the rest length of edge $(i,j)$,  
and $\theta_{pq}^0$ is the rest dihedral angle at hinge $(p,q)$. The material parameters are selected to approximate the properties of the nitrile sheet.  
The surface density $\rho_s$ is chosen so that the simulated total mass matches the physical prototype.  
The stiffness coefficients $(k_s, k_b)$ are tuned to balance container conformity with sufficient structural integrity. The damping coefficient $c$ is introduced to suppress spurious oscillations during simulation. In practice, seventy-two peripheral vertices are exposed and coupled to colocated spherical geoms via equality \emph{weld} constraints, which act as proxies for contact and actuation while preserving the compliance of the underlying deformable model.

\paragraph{Control}  
To enable controllable deformation, the sheet boundary is discretized into vertices, each associated with a small proxy geom. As illustrated in the bottom row of Fig.~\ref{fig: sim_scu}, based on observations from the physical prototype, we establish mapping rules that determine which vertex pairs should be linked to reproduce a given funnel angle. This mapping captures the correspondence between geometric connections and the resulting global deformation. For robustness, the activation of these constraints is scheduled in stages. This staged strategy ensures smooth transitions, prevents over-constraining, and yields reliable control of the deformable sheet while preserving its compliance.

\subsection{Problem Formulation}
\label{subsec:problem_formulation}
The objective is to enable a soft, deformable robotic hand to perform scooping across diverse containers, adapting its shape and size much like humans intuitively select tools. To address this, we construct consistent simulation environments and formulate the task as an optimization problem in trajectory generation under interaction dynamics, where both the motion and the hand’s rolling angle are decision variables. All environments are generated from real-world setups into a simulation, making our framework akin to a digital twin system. This allows the optimized trajectories to transfer to real robots with minimal sim-to-real gap.

\begin{figure*}[!t]
      \centering
      \includegraphics[width=1.0\linewidth]{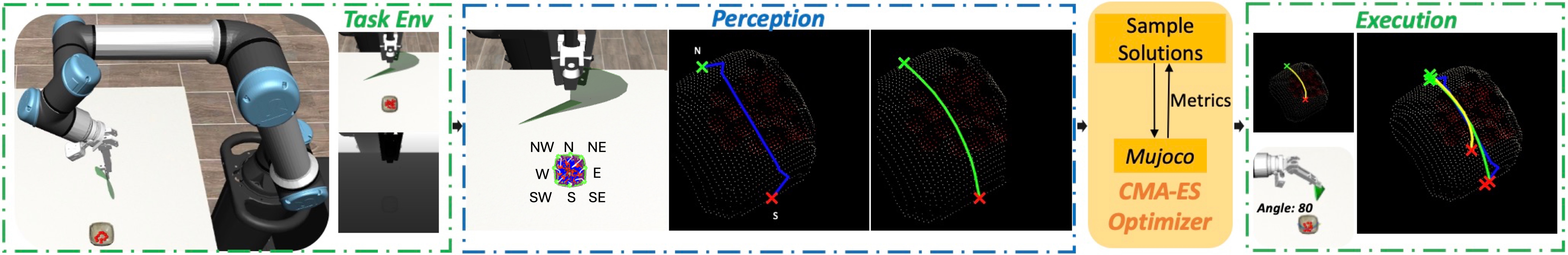}
      \vspace{-4mm}
      \caption{Overall framework of our system: from RGB–D perception and ring–compass abstraction, through seed trajectory generation and parameterization, to trajectory optimization with CMA-ES, enabling transfer of optimized scooping strategies from simulation to the real robot.}
      \label{fig: framework}
\end{figure*}

\subsubsection{Observation}
The input is an RGB-D image $\mathcal{I}$, from which we extract container keypoints
$\mathcal{P} = \{\mathbf{p}_i \in \mathbb{R}^3 \mid i = 0, \dots, M\}$ where $M$ is the number of keypoints and construct a seed path that connects anchor points (e.g., \texttt{topN} $\rightarrow$ center $\rightarrow$ \texttt{topS}), where top denotes on the container’s top rim. Illustrated in Fig.~\ref{fig: framework}.

\subsubsection{Scooping Action Parameterization}
We aim to optimize the trajectory and the rolling angle of the SH jointly.

\begin{figure}[!t]
      \centering
      \includegraphics[width=1.0\linewidth]{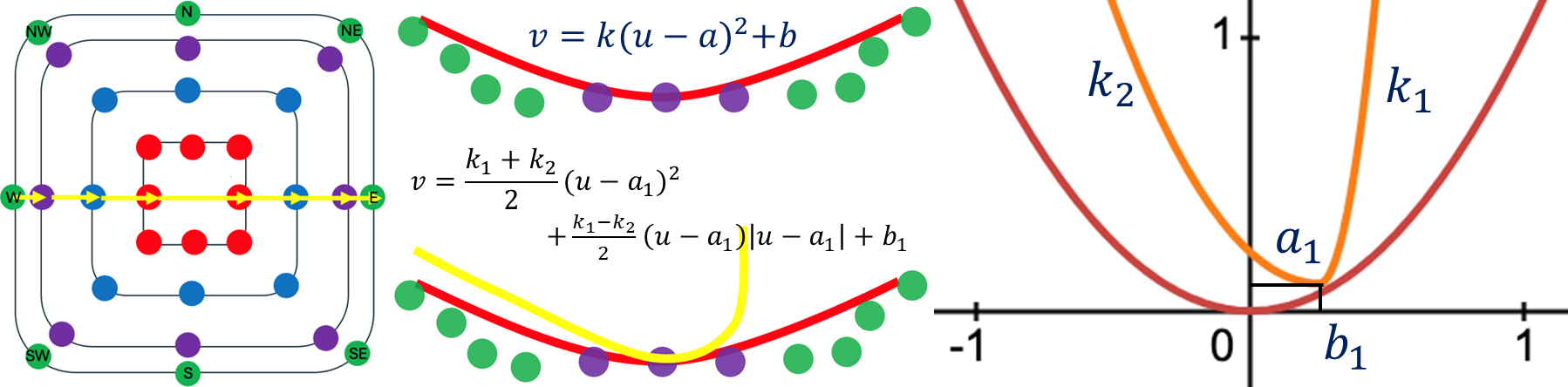}

      \caption{From left to right: anchor skeleton, parabolic fit in uv-plane, and piecewise-parabola trajectory.}
      \label{fig: opt_traj_math}
\end{figure}

\paragraph{Scoop Trajectory Analysis}
Fig.~\ref{fig: opt_traj_math} shows that we fit a local motion plane using singular value decomposition (SVD), which is equivalent to principal component analysis (PCA) on the point set. This defines a right-handed coordinate frame $(\mathbf{e}_1,\mathbf{e}_2,\mathbf{n})$ with basis vectors spanning the plane and a normal vector $\mathbf{n}$. Any 3D point can then be projected into local coordinates $(u,v)$.  

The three anchors (\texttt{topN}, center, \texttt{topS}) are projected into the local plane and used to fit a parabola
\begin{equation}
v = A u^2 + B u + C
\end{equation}
We convert it to vertex form
\begin{equation}
v = k(u-a)^2 + b, \qquad
k=A, \quad a=-\tfrac{B}{2A}, \quad b=C - Aa^2
\end{equation}
and denote the seed by parameters $(k,a,b)$. The endpoints $(u_{\text{start}},v_{\text{start}})$ and $(u_{\text{end}},v_{\text{end}})$ are also recorded.  
 
To allow more flexibility, the optimized trajectory is modeled as a continuous, piecewise parabola with potentially different curvatures on either side of the vertex:
\begin{equation}
\begin{split}
v(u; k_1,k_2,a_1,b_1) = &
\frac{k_1+k_2}{2}(u-a_1)^2 \\
&+ \frac{k_1-k_2}{2}(u-a_1)\lvert u-a_1\rvert + b_1
\end{split}
\end{equation}

This ensures $C^1$ continuity while allowing asymmetric curvature. $\bar{\mathbf{p}}$ denotes the reference point in world coordinates. The world-space curve is then obtained by mapping back via
\begin{equation}
\mathbf{x}(u) = \bar{\mathbf{p}} + u\,\mathbf{e}_1 + v(u)\,\mathbf{e}_2 
\end{equation}

Since our scooping task always takes place in the x–z or y–z plane, constraining the z-position is meaningful. Therefore, the curve is trimmed so that its endpoints align with the seed anchors, and it is then resampled into $n$ evenly spaced waypoints along its arc length, which provides a consistent parameterization for optimization and execution.

\paragraph{Soft Hand Roll Angle and Rotation of End-effector}

From the container observation, we obtain an initial estimate of the SH roll angle $angle_{initial}$ related to the size of the container. This value serves as an initial angle, while the optimizer explores variations by adding a continuous offset $\Delta angle$. The optimizer evaluates candidate roll angles by interacting with the MuJoCo simulation environment, enabling identification of the most effective roll angle for the scooping task. Since the roll angle directly affects the orientation of the sheet funnel’s edge relative to the end effector (Fig.~\ref{fig: sim_scu}), to maintain a consistent sheet orientation relative to the ground, the optimized roll angle is remapped into an appropriate rotation range for the end effector. This ensures that both insertion into and withdrawal from the container remain feasible.

For consecutive waypoints $(\mathbf{x}_i,\mathbf{x}_{i+1})$, we compute the motion direction to define a tilt angle $tiltdeg_i$. Pitch of the end-effector is then obtained as a function of $tiltdeg_i$ and the discrete SH roll angle $angle_i$:
\begin{equation}
pitch_i \;=\; m(tiltdeg_i;angle_i)
\end{equation}
where $m(\cdot;\cdot)$ is the piecewise-linear band mapping.  

Fig.~\ref{fig: sim_scu} shows that six discrete roll angles are defined for the SH. Specifically, the roll angle is discretized into buckets:
\begin{equation}
\mathcal{A} = \{20^\circ, 40^\circ, 60^\circ, 80^\circ, 100^\circ, 120^\circ\}
\end{equation}

\begin{equation}
\tilde{a}_i = \min\!\bigl(\max(angle_{{initial}} + \Delta angle,\,20^\circ),\,120^\circ\bigr)
\end{equation}

\begin{equation}
angle_i = \operatorname*{argmin}_{a \in \mathcal{A}} \; \lvert a - \tilde{a}_i \rvert
\end{equation}

where $angle_{initial}$ denotes the prior roll angle and $\Delta angle$ is the decision variable determined by the optimizer.

\subsubsection{Evaluation Metrics}

As the optimizer interacts with the simulation at each trial, we define an objective function based on three key aspects: the volume of material successfully scooped, the amount of overflow spilled outside the container, and whether unintended collisions occur between the hand and the container. These criteria jointly capture both efficiency and feasibility of the scooping motion.

\subsection{Scooping Trajectory Generation}
\label{subsec:scoop_gen} 

\subsubsection{Visual Perception}  
\begin{figure}[!t]
      \centering
      \includegraphics[width=1.0\linewidth]{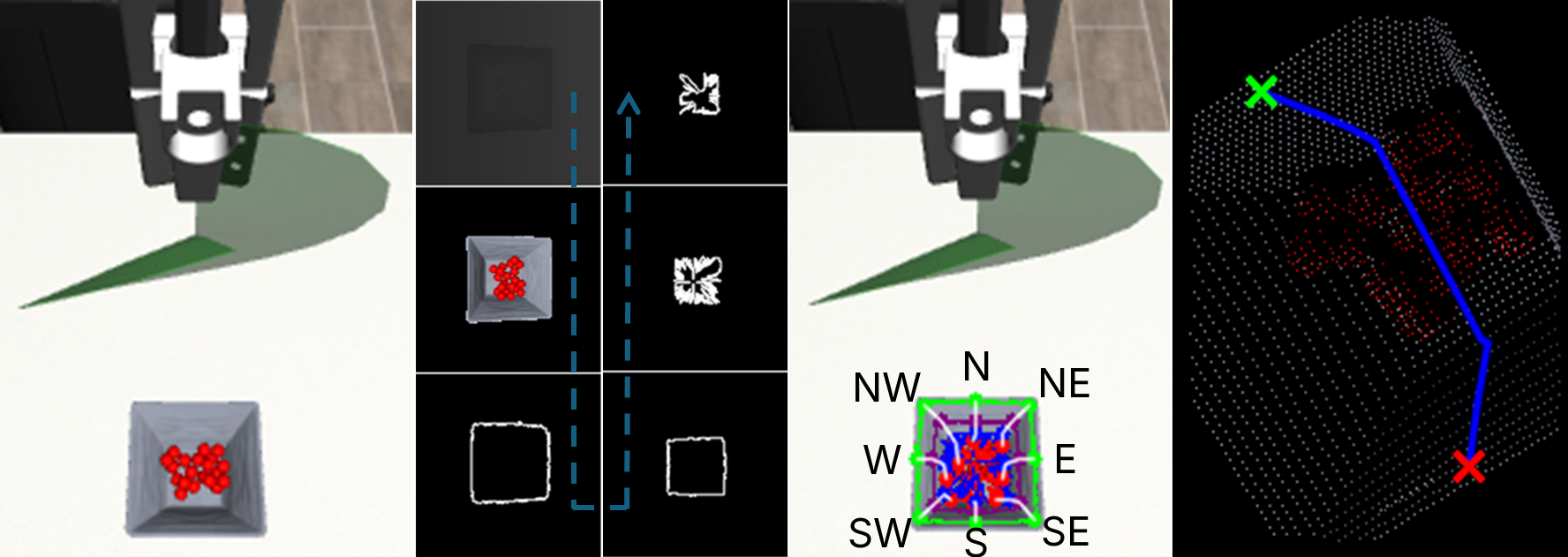}
       
      \caption{Perception pipeline: RGB–D input to ring–compass representation for trajectory planning.}
      \label{fig: perception}
\end{figure}

To initialize scooping, the system first detects the target container and estimates its geometry. Using FastSAM \cite{zhao2023fast}, object masks are extracted from RGB images and fused with depth data to reconstruct a partial height field in world coordinates. This field is vertically bounded by observed heights, and several iso-height surfaces are defined to approximate the container’s functional regions: base, walls, and rim. At each iso-height level, we extract the closed contour formed by the mask–height band intersection. From each contour, eight equi-angular “compass” points are sampled with respect to the centroid, yielding a structured set of control points ${p_{r,d}}$, where $r$ indexes the ring level and $d$ the directional label. Connecting points with identical directions across levels produces eight vertical profiles that approximate the container’s inner surface from base to rim (see Fig.~\ref{fig: perception}). This ring–compass representation offers a compact geometric abstraction of the container, capturing both global radial symmetry and local variations in rim curvature and wall inclination. These features act as geometric priors for trajectory planning, guiding insertion and deformation control of the hand.

\subsubsection{Generation of Initial Trajectories}  

\begin{figure}[!t]
      \centering
      \includegraphics[width=1.0\linewidth]{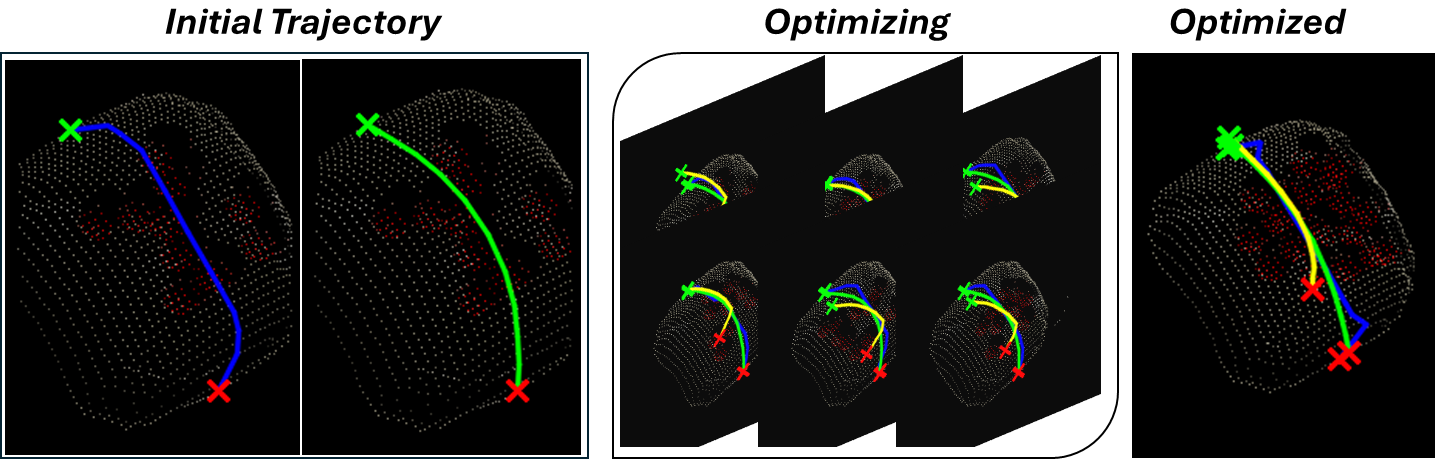}

      \caption{Generation and refinement of scooping trajectories. From left to right: initial trajectory seeded from the ring–compass skeleton, intermediate optimization process, and final optimized trajectory.}
      \label{fig: opt_traj}
\end{figure}

Given the ring--compass skeleton $\{p_{r,d}\}$ extracted from perception, we generate a parameterized waypoint sequence to guide the end-effector from free space to insertion and lift-out. As shown in Fig.~\ref{fig: opt_traj}, a compass direction $d^\star \in \{N, NE, E, SE, S, SW, W, NW\}$ is first selected based on task heuristics such as approach side, occlusion, or arm reachability. Along this direction, the vertical profile
\[
\mathcal{P}_{d^\star}=\big[p_{\mathrm{bottom},d^\star},\;p_{\mathrm{mid1},d^\star},\;p_{\mathrm{mid2},d^\star},\;p_{\mathrm{top},d^\star}\big]
\]
provides four seed points from base to rim, which are connected into a piecewise linear path. Each waypoint is assigned an end-effector orientation, initialized as described in Sec.~\ref{subsec:problem_formulation}. The initial angle of the deformable hand is set according to the container size. Following Sec.~\ref{subsec:problem_formulation}, the scooping trajectory is formulated as an optimization problem.

\subsubsection{Trajectory Optimization via CMA-ES}
CMA-ES is a robust evolutionary strategy for real-valued optimization in continuous domains, well-suited to stochastic, nonlinear, and nonconvex functions \cite{hansen1996adapting}. It maintains a population of candidate solutions and updates them through selection, recombination, and mutation. As described in Sec.~\ref{subsec:problem_formulation}, our initial trajectory is parameterized, requiring optimization of four parameters to shape the curve and one additional parameter to set the hand roll.

\paragraph{Parameterization}  
We optimize five variables: four curve parameters and the hand roll angle. The optimization vector is
\begin{equation}
\label{eq:opt_vector}
\theta = \big(k_1,\, k_2,\, a_1,\, b_1,\, \Delta angle\big)
\end{equation}

Using the single-parabola seed $(k,a,b)$, we define data-driven bounds:
\begin{align}
k_1 &\in \operatorname{sign}(k)\cdot\bigl(2|k|,\, 20|k|\bigr], 
&& \label{eq:k1_bound} \\
k_2 &\in \operatorname{sign}(k)\cdot\bigl(0.8|k|,\, 1.2|k|\bigr], 
&& \label{eq:k2_bound} \\
a_1 &\in [a_{\mathrm{lo}},\, a_{\mathrm{hi}}], 
&& \label{eq:a1_bound} \\
b_1 &\in \bigl(b,\, b+\Delta b\bigr], \qquad \Delta b \approx 0.01, 
&& \label{eq:b1_bound} \\
\Delta\!\text{angle} &\in [-\Delta_{\max},\, \Delta_{\max}], 
\qquad \Delta_{\max} \approx 40^\circ 
&& \label{eq:angle_bound}
\end{align}
where $a_{\mathrm{lo}}$ and $a_{\mathrm{hi}}$ correspond to the two waypoints before and after the center waypoint of the initial trajectory.

\paragraph{Objective Function} 
Each candidate trajectory is simulated in MuJoCo with the SH deformable model at a chosen discrete opening $\theta$. 
During rollout, task metrics are extracted:  
(i) \emph{scooped} — number of particles lifted above a container-specific $Z$ threshold;  
(ii) \emph{overflow} — particles remaining below threshold;  
(iii) \emph{collision} — indicator if any robot body contacts the container or table.  

The scalar loss is defined as
\begin{equation}
\label{eq:loss}
\begin{split}
\mathcal{L}(\vartheta,\theta) =\;&
-w_1\,\mathrm{scooped} + w_2\,\mathrm{overflow} \\
&+ w_3\,\mathrm{collision} + \lambda\,\mathbf{I}
\end{split}
\end{equation}
where $\mathbf{I}$ penalizes trajectories that fail to reach all waypoints.  

CMA-ES samples and rebuilds the trajectory, and evaluates $\mathcal{L}$ in simulation. The optimizer returns the best parameters and trajectory, together with diagnostics such as per-term metrics, consistently improving scooping yield while reducing overflow and collisions.

\section{Experiments}
\label{sec:experiments}

\subsection{Experimental Setups}
\label{subsec:exp_setups}

\begin{figure}[!t]
      \centering
      \includegraphics[width=.9\linewidth]{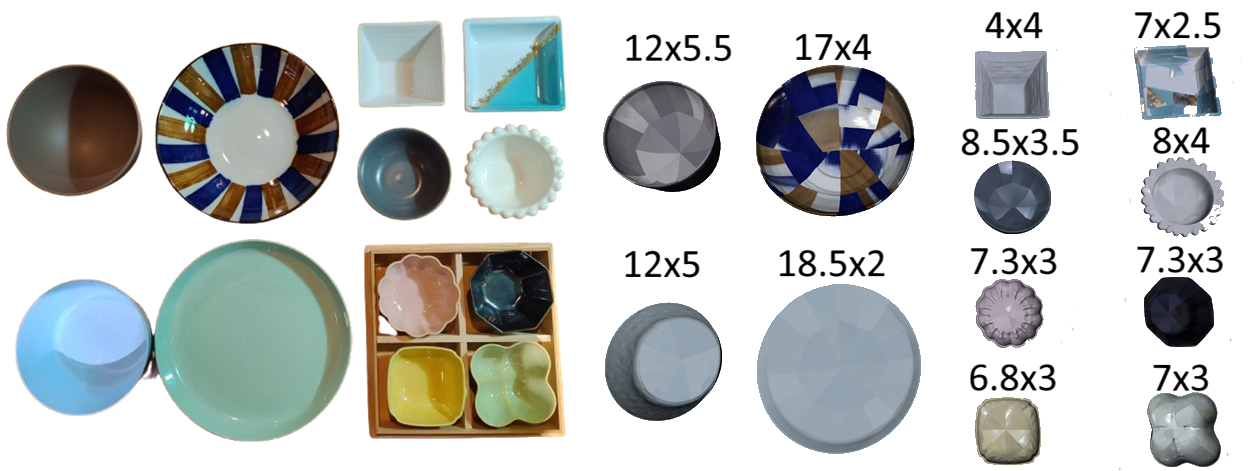}
      \par\noindent\makebox[0.48\linewidth][c]{\footnotesize Real-world Objects} 
      \makebox[0.48\linewidth][c]{\footnotesize Simulation Objects (cm)}

      \caption{Containers used in experiments: $12$ objects in three sizes (large, medium, small), including regular and irregular shapes. (Width x Height)}
      \label{fig: objects}
\end{figure}

As illustrated in Fig.~\ref{fig: overview}, the left shows the real robot and the right the corresponding simulation environment, designed for repeatable scooping and camera observation. Fig.~\ref{fig: objects} shows that experiments use granular media in containers of varying shapes and sizes (bowls and plates, circular, square, and irregular). The objective is to assess the fidelity of the hand simulation: whether it reproduces the deformation and interaction of the real SH, and whether optimized trajectories transfer with minimal sim-to-real gap. We conducted two types of experiments. First, ablation studies examined which parameters should be optimized. Second, we evaluated the success rate of trajectories optimized in simulation and transferred them to the real robot.

\subsection{Simulation Experiments}
\label{subsec:sim_exp}

\subsubsection{Ablation Experiments}
\label{subsubsec:ablation_exp}
Scooping performance is influenced by multiple trajectory model parameters. To assess their relative contributions, we performed ablation experiments in simulation with a standard bowl (Table~\ref{table: ablation}). In each variant, one parameter was excluded from optimization: $k_1$, $k_2$, $a_1$, $b_1$, or $\Delta angle$. The results highlight that curvature parameters ($k_1$, $k_2$) are the most critical: removing either sharply reduces performance, with success counts dropping to nearly half of the full optimization setting. By contrast, excluding vertex parameters ($a_1$, $b_1$) causes smaller but still noticeable degradation. Excluding the rotation term $\Delta angle$ also reduces scooping efficiency, particularly for small containers. Overall, these results confirm that robust scooping requires jointly tuning all parameters curvature, vertex, and rotation underscoring the importance of full optimization for consistent performance across container sizes.

\begin{table}[!t]
    \centering
    \caption{Scooping results under different action parameterizations across 3 container sizes in the 10 balls task.}
    \resizebox{\linewidth}{!}{
    \begin{tabular}{l c c c c}
        \toprule
        \multirow{2}{*}{Action Parameterization} & \multirow{2}{*}{Parameters} & \multicolumn{3}{c}{Scooped} \\
        \cline{3-5}
         &  & Big & Medium & Small \\
        \midrule
        Initialization     & $(k,a,b,\Delta angle)$                 & 0 & 0 & 0 \\
        No $k_1$           & $(k_2,a_1,b_1,\Delta angle)$           & 2 & 1 & 1 \\
        No $k_2$           & $(k_1,a_1,b_1,\Delta angle)$           & 6 & 5 & 5 \\
        No $a_1$           & $(k_1,k_2,b_1,\Delta angle)$           & 6 & 7 & 8 \\
        No $b_1$           & $(k_1,k_2,a_1,\Delta angle)$           & 8 & 9 & 9 \\
        No $\Delta angle$           & $(k_1,k_2,a_1,b_1)$          &  8 & 8  & 6  \\
        Full optimization  & $(k_1,k_2,a_1,b_1,\Delta angle)$       & 10 & 10 & 9 \\
        \bottomrule
    \end{tabular}}
    \label{table: ablation}
\end{table}

\subsection{Real-World Experiments}
\label{subsec:real}
\begin{figure*}[!t]
      \centering
      \includegraphics[width=0.49\linewidth]{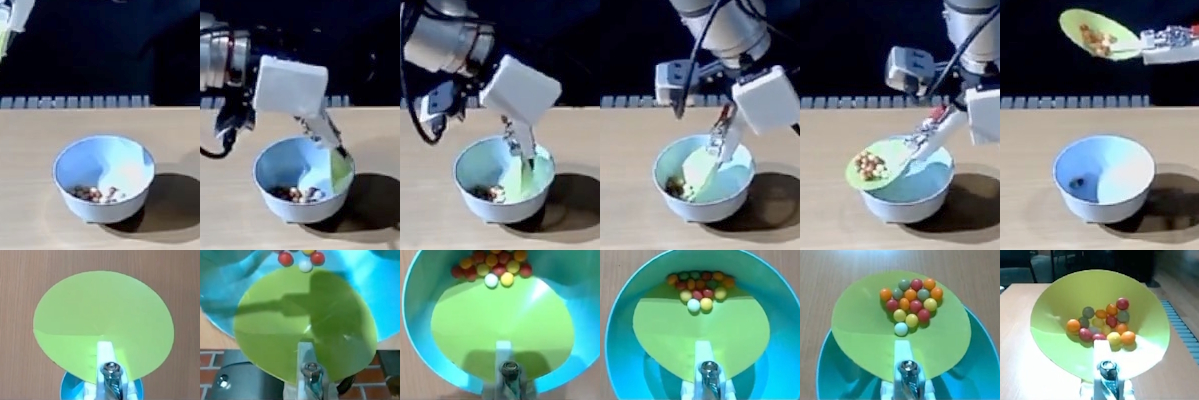}
      \includegraphics[width=0.49\linewidth]{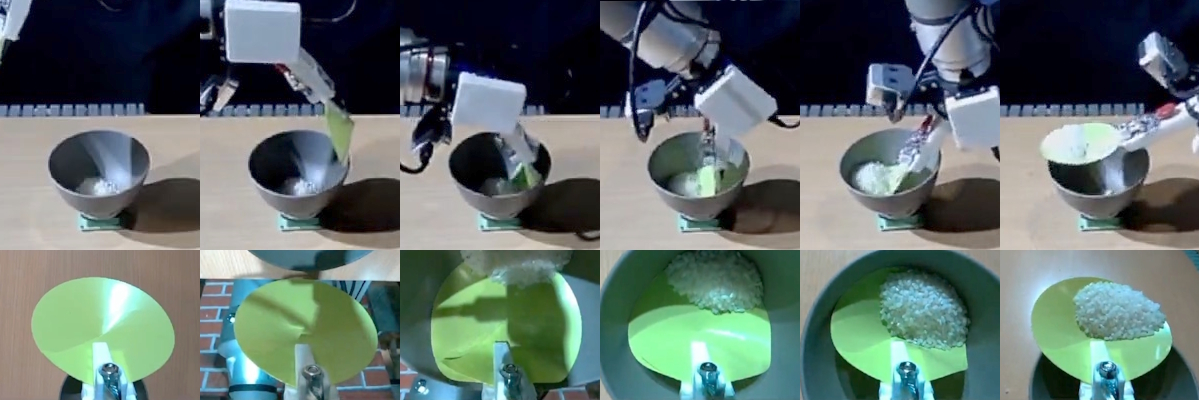}
      \vspace{-4mm}
      \caption{Action sequence of the SH executing the optimized trajectory transferred from simulation to reality. The figure illustrates key phases of the scooping process, including rolling SH, insertion,  and lifting. The left is the 20-ball task, and the right is the 50g-rice task.}
      \label{fig: scoop_actions_real}
\end{figure*}

We optimized trajectories exclusively in simulation on the 10-ball task, where the best solution achieved a perfect outcome, scooping all ten balls. This single optimized trajectory was then directly transferred to the real SH system without any further tuning. Figures~\ref{fig: overview} and \ref{fig: scoop_actions_real} illustrate the experimental setup and action sequence, while Table~\ref{table: tasks_trials} reports the quantitative results. In real experiments, the robot scooped on average $7.9$ balls in the 10-ball task, $14.9$ in the 20-ball task, and $48.25$ g in the rice task. Although real-world performance showed some variability due to sensing and actuation noise, one trial in the 10-ball task exactly matched the simulation result, successfully scooping all ten balls. Importantly, despite being optimized only on the 10-ball task, the same trajectory generalized effectively to the 20-ball and rice tasks, demonstrating strong zero-shot transfer across object types and container conditions. These findings confirm that the digital twin framework enables robust sim-to-real transfer: the deformable hand model faithfully reproduces the key behaviors of the physical system, and trajectories optimized in simulation generalize to real execution with minimal adaptation. Supplementary videos provide additional evidence, illustrating both the scooping process and the high fidelity of the transferred strategies.

\begin{table*}[!t]
  \centering
  \caption{Scooping results over 10 trials for three tasks in a real robot.}
  \resizebox{\linewidth}{!}{
  \begin{tabular}{lccccccccccc}
    \toprule
    \textbf{Task} & \textbf{Trial 1} & \textbf{2} & \textbf{3} & \textbf{4} & \textbf{5} & 
                   \textbf{6} & \textbf{7} & \textbf{8} & \textbf{9} & \textbf{10} & \textbf{Average} \\
    \midrule
    10-balls      & 10 & 7  & 8  & 8  & 5  & 9  & 5  & 9  & 9  & 9  & \textbf{7.9} \\
    20-balls      & 16 & 16 & 16 & 13 & 15 & 15 & 14 & 14 & 14 & 16 & \textbf{14.9} \\
    50g-rice (g)  & 48.23 & 47.99 & 48.12 & 48.27 & 47.82 & 47.92 & 49.06 & 48.13 & 48.60 & 48.38 & \textbf{48.25} \\
    \bottomrule
  \end{tabular}}
  \label{table: tasks_trials}
\end{table*}

\section{Conclusion}
\label{sec:conclusion}

This paper presents a comprehensive framework that integrates an existing soft, deformable conical robotic hand with a physics-based MuJoCo simulation model to enable robust, adaptive scooping across varied containers. The hand’s passive compliance enables it to conform to various container geometries, while the physics-based simulation accurately reproduces its deformation and interaction with granular materials. Visual perception is utilized to generate initial scooping trajectories that are then optimized within the simulation environment using the CMA-ES evolutionary strategy. The resulting optimized trajectories transfer effectively to the physical robot with minimal sim-to-real gap. 

This approach addresses the limitations of rigid and hand-engineered methods by leveraging soft tools and simulation-driven trajectory optimization, revealing significant potential for advancing robot-assisted manipulation. Future work will explore extending the framework to more complex materials and incorporating learning-based trajectory optimization for broader applications.

\bibliographystyle{ieeetr}
\bibliography{reference}

@article{yang2025s,
  title={S$^2$-Diffusion: Generalizing from Instance-level to Category-level Skills in Robot Manipulation},
  author={Yang, Quantao and Welle, Michael C and Kragic, Danica and Andersson, Olov},
  journal={arXiv preprint arXiv:2502.09389},
  year={2025}
}

@article{franceschini2024autonomous,
  title={Autonomous Excavation of Challenging Terrain using Oscillatory Primitives and Adaptive Impedance Control},
  author={Franceschini, Noah and Thangeda, Pranay and Ornik, Melkior and Hauser, Kris},
  journal={arXiv preprint arXiv:2409.18273},
  year={2024}
}

@inproceedings{tai2023scone,
  title={Scone: A food scooping robot learning framework with active perception},
  author={Tai, Yen-Ling and Chiu, Yu Chien and Chao, Yu-Wei and Chen, Yi-Ting},
  booktitle={Conference on Robot Learning},
  pages={849--865},
  year={2023},
  organization={PMLR}
}

@inproceedings{niu2023goats,
  title={Goats: Goal sampling adaptation for scooping with curriculum reinforcement learning},
  author={Niu, Yaru and Jin, Shiyu and Zhang, Zeqing and Zhu, Jiacheng and Zhao, Ding and Zhang, Liangjun},
  booktitle={2023 IEEE/RSJ International Conference on Intelligent Robots and Systems (IROS)},
  pages={1023--1030},
  year={2023},
  organization={IEEE}
}

@article{lin2025robotsmith,
  title={RobotSmith: Generative Robotic Tool Design for Acquisition of Complex Manipulation Skills},
  author={Lin, Chunru and Yuan, Haotian and Wang, Yian and Qiu, Xiaowen and Wang, Tsun-Hsuan and Guo, Minghao and Wang, Bohan and Narang, Yashraj and Fox, Dieter and Gan, Chuang},
  journal={arXiv preprint arXiv:2506.14763},
  year={2025}
}

@article{mehta2025l2d2,
  title={L2D2: Robot Learning from 2D Drawings},
  author={Mehta, Shaunak A and Nemlekar, Heramb and Sumant, Hari and Losey, Dylan P},
  journal={arXiv preprint arXiv:2505.12072},
  year={2025}
}

@inproceedings{kageyama2024learning,
  title={Learning Scooping Deformable Plastic Objects using Tactile Sensors},
  author={Kageyama, Yuto and Hamaya, Masashi and Tanaka, Kazutoshi and Hashimoto, Atsushi and Saito, Hideo},
  booktitle={2024 IEEE 20th International Conference on Automation Science and Engineering (CASE)},
  pages={4020--4025},
  year={2024},
  organization={IEEE}
}

@inproceedings{schenck2017learning,
  title={Learning robotic manipulation of granular media},
  author={Schenck, Connor and Tompson, Jonathan and Levine, Sergey and Fox, Dieter},
  booktitle={Conference on Robot Learning},
  pages={239--248},
  year={2017},
  organization={PMLR}
}

@inproceedings{song2025soda,
  title={SODA--Soft Origami Dynamic Utensil for Assisted Feeding},
  author={Song, Yuxin Ray and Luo, Yiyue},
  booktitle={2025 IEEE 8th International Conference on Soft Robotics (RoboSoft)},
  pages={1--8},
  year={2025},
  organization={IEEE}
}

@inproceedings{franco2024double,
  title={The double-scoop gripper: A tendon-driven soft-rigid end-effector for food handling exploiting constraints in narrow spaces},
  author={Franco, Leonardo and Turco, Enrico and Bo, Valerio and Pozzi, Maria and Malvezzi, Monica and Prattichizzo, Domenico and Salvietti, Gionata},
  booktitle={2024 IEEE International Conference on Robotics and Automation (ICRA)},
  pages={4170--4176},
  year={2024},
  organization={IEEE}
}

@article{wang2021learning,
  title={Learning compositional models of robot skills for task and motion planning},
  author={Wang, Zi and Garrett, Caelan Reed and Kaelbling, Leslie Pack and Lozano-P{\'e}rez, Tom{\'a}s},
  journal={The International Journal of Robotics Research},
  volume={40},
  number={6-7},
  pages={866--894},
  year={2021},
  publisher={SAGE Publications Sage UK: London, England}
}

@inproceedings{liu2024diff,
  title={Diff-Control: A Stateful Diffusion-based Policy for Imitation Learning},
  author={Liu, Xiao and Zhou, Yifan and Weigend, Fabian and Sonawani, Shubham and Ikemoto, Shuhei and Amor, Heni Ben},
  booktitle={2024 IEEE/RSJ International Conference on Intelligent Robots and Systems (IROS)},
  pages={7453--7460},
  year={2024},
  organization={IEEE}
}

@article{huang2025dih,
  title={Dih-tele: Dexterous in-hand teleoperation framework for learning multiobjects manipulation with tactile sensing},
  author={Huang, Junda and Chen, Kai and Zhou, Jianshu and Lin, Xingyu and Abbeel, Pieter and Dou, Qi and Liu, Yunhui},
  journal={IEEE/ASME Transactions on Mechatronics},
  year={2025},
  publisher={IEEE}
}

@article{moorman2024investigating,
  title={Investigating strategies enabling novice users to teach plannable hierarchical tasks to robots},
  author={Moorman, Nina and Singh, Aman and Natarajan, Manisha and Hedlund-Botti, Erin and Schrum, Mariah and Yang, Chuxuan and Seelam, Lakshmi and Gombolay, Matthew C and Gopalan, Nakul},
  journal={The International Journal of Robotics Research},
  pages={02783649241301075},
  year={2024},
  publisher={SAGE Publications Sage UK: London, England}
}

@article{keely2025kiri,
  title={Kiri-Spoon: A Kirigami Utensil for Robot-Assisted Feeding},
  author={Keely, Maya and Franco, Brandon and Grothoff, Casey and Jenamani, Rajat Kumar and Bhattacharjee, Tapomayukh and Losey, Dylan P and Nemlekar, Heramb},
  journal={arXiv preprint arXiv:2501.01323},
  year={2025}
}

@inproceedings{bhaskar2024lava,
  title={Lava: Long-horizon visual action based food acquisition},
  author={Bhaskar, Amisha and Liu, Rui and Sharma, Vishnu D and Shi, Guangyao and Tokekar, Pratap},
  booktitle={2024 IEEE/RSJ International Conference on Intelligent Robots and Systems (IROS)},
  pages={8929--8935},
  year={2024},
  organization={IEEE}
}

@article{ruan2024primp,
  title={PRIMP: Probabilistically-informed motion primitives for efficient affordance learning from demonstration},
  author={Ruan, Sipu and Liu, Weixiao and Wang, Xiaoli and Meng, Xin and Chirikjian, Gregory S},
  journal={IEEE Transactions on Robotics},
  volume={40},
  pages={2868--2887},
  year={2024},
  publisher={IEEE}
}

@article{zhao2025learning,
  title={Learning thin deformable object manipulation with a multi-sensory integrated soft hand},
  author={Zhao, Chao and Jiang, Chunli and Luo, Lifan and Yuan, Shuai and Chen, Qifeng and Yu, Hongyu},
  journal={IEEE Transactions on Robotics},
  year={2025},
  publisher={IEEE}
}

@article{das2024screw,
  title={Screw Geometry Meets Bandits: Incremental Acquisition of Demonstrations to Generate Manipulation Plans},
  author={Das, Dibyendu and Patankar, Aditya and Chakraborty, Nilanjan and Ramakrishnan, CR and Ramakrishnan, IV},
  journal={arXiv preprint arXiv:2410.18275},
  year={2024}
}

@article{zhu2022soft,
  title={A soft-rigid hybrid gripper with lateral compliance and dexterous in-hand manipulation},
  author={Zhu, Wenpei and Lu, Chenghua and Zheng, Qule and Fang, Zhonggui and Che, Haichuan and Tang, Kailuan and Zhu, Mingchao and Liu, Sicong and Wang, Zheng},
  journal={IEEE/ASME Transactions on Mechatronics},
  volume={28},
  number={1},
  pages={104--115},
  year={2022},
  publisher={IEEE}
}

@inproceedings{mehta2023riso,
  title={RISO: Combining rigid grippers with soft switchable adhesives},
  author={Mehta, Shaunak A and Kim, Yeunhee and Hoegerman, Joshua and Bartlett, Michael D and Losey, Dylan P},
  booktitle={2023 IEEE International Conference on Soft Robotics (RoboSoft)},
  pages={1--8},
  year={2023},
  organization={IEEE}
}

@inproceedings{wang2025dexgrip,
  title={DexGrip: Multi-modal Soft Gripper with Dexterous Grasping and In-hand Manipulation Capacity},
  author={Wang, Xing and Horrigan, Liam and Pinskier, Josh and Shi, Ge and Viswanathan, Vinoth and Liow, Lois and Bandyopadhyay, Tirthankar and Chung, Jen Jen and Howard, David},
  booktitle={2025 IEEE 8th International Conference on Soft Robotics (RoboSoft)},
  pages={1--6},
  year={2025},
  organization={IEEE}
}

@article{guo2025fish,
  title={Fish Mouth Inspired Origami Gripper for Robust Multi-Type Underwater Grasping},
  author={Guo, Honghao and Huang, Junda and Zhang, Ian and Liang, Boyuan and Ma, Xin and Liu, Yunhui and Zhou, Jianshu},
  journal={arXiv preprint arXiv:2503.11049},
  year={2025}
}

@article{takahashi2025scu,
  title={SCU-Hand: Soft Conical Universal Robotic Hand for Scooping Granular Media from Containers of Various Sizes},
  author={Takahashi, Tomoya and Beltran-Hernandez, Cristian C and Kuroda, Yuki and Tanaka, Kazutoshi and Hamaya, Masashi and Ushiku, Yoshitaka},
  journal={arXiv preprint arXiv:2505.04162},
  year={2025}
}

@inproceedings{todorov2012mujoco,
  title={Mujoco: A physics engine for model-based control},
  author={Todorov, Emanuel and Erez, Tom and Tassa, Yuval},
  booktitle={2012 IEEE/RSJ international conference on intelligent robots and systems},
  pages={5026--5033},
  year={2012},
  organization={IEEE}
}

@article{zakka2025mujoco,
  title={Mujoco playground},
  author={Zakka, Kevin and Tabanpour, Baruch and Liao, Qiayuan and Haiderbhai, Mustafa and Holt, Samuel and Luo, Jing Yuan and Allshire, Arthur and Frey, Erik and Sreenath, Koushil and Kahrs, Lueder A and others},
  journal={arXiv preprint arXiv:2502.08844},
  year={2025}
}

@inproceedings{hansen1996adapting,
  title={Adapting arbitrary normal mutation distributions in evolution strategies: The covariance matrix adaptation},
  author={Hansen, Nikolaus and Ostermeier, Andreas},
  booktitle={Proceedings of IEEE international conference on evolutionary computation},
  pages={312--317},
  year={1996},
  organization={IEEE}
}

@article{bacher2021design,
  title={Design and control of soft robots using differentiable simulation},
  author={B{\"a}cher, Moritz and Knoop, Espen and Schumacher, Christian},
  journal={Current Robotics Reports},
  volume={2},
  number={2},
  pages={211--221},
  year={2021},
  publisher={Springer}
}

@inproceedings{lin2021softgym,
  title={Softgym: Benchmarking deep reinforcement learning for deformable object manipulation},
  author={Lin, Xingyu and Wang, Yufei and Olkin, Jake and Held, David},
  booktitle={Conference on Robot Learning},
  pages={432--448},
  year={2021},
  organization={PMLR}
}

@article{naughton2021elastica,
  title={Elastica: A compliant mechanics environment for soft robotic control},
  author={Naughton, Noel and Sun, Jiarui and Tekinalp, Arman and Parthasarathy, Tejaswin and Chowdhary, Girish and Gazzola, Mattia},
  journal={IEEE Robotics and Automation Letters},
  volume={6},
  number={2},
  pages={3389--3396},
  year={2021},
  publisher={IEEE}
}

@inproceedings{fuchioka2024robotic,
  title={Robotic Object Insertion with a Soft Wrist through Sim-to-Real Privileged Training},
  author={Fuchioka, Yuni and Beltran-Hernandez, Cristian C and Nguyen, Hai and Hamaya, Masashi},
  booktitle={2024 IEEE/RSJ International Conference on Intelligent Robots and Systems (IROS)},
  pages={9159--9166},
  year={2024},
  organization={IEEE}
}

@inproceedings{peng20243d,
  title={3d force and contact estimation for a soft-bubble visuotactile sensor using fem},
  author={Peng, Jing-Chen and Yao, Shaoxiong and Hauser, Kris},
  booktitle={2024 IEEE International Conference on Robotics and Automation (ICRA)},
  pages={5666--5672},
  year={2024},
  organization={IEEE}
}

@inproceedings{xu2023efficient,
  title={Efficient tactile simulation with differentiability for robotic manipulation},
  author={Xu, Jie and Kim, Sangwoon and Chen, Tao and Garcia, Alberto Rodriguez and Agrawal, Pulkit and Matusik, Wojciech and Sueda, Shinjiro},
  booktitle={Conference on Robot Learning},
  pages={1488--1498},
  year={2023},
  organization={PMLR}
}

@article{zhu2022review,
  title={A review on origami simulations: from kinematics, to mechanics, toward multiphysics},
  author={Zhu, Yi and Schenk, Mark and Filipov, Evgueni T},
  journal={Applied Mechanics Reviews},
  volume={74},
  number={3},
  pages={030801},
  year={2022},
  publisher={American Society of Mechanical Engineers}
}

@article{tachi2009simulation,
  title={Simulation of rigid origami},
  author={Tachi, Tomohiro},
  journal={Origami},
  volume={4},
  number={08},
  pages={175--187},
  year={2009},
  publisher={Kerswell Books}
}

@article{zhao2023fast,
  title={Fast segment anything},
  author={Zhao, Xu and Ding, Wenchao and An, Yongqi and Du, Yinglong and Yu, Tao and Li, Min and Tang, Ming and Wang, Jinqiao},
  journal={arXiv preprint arXiv:2306.12156},
  year={2023}
}

\end{document}